\documentclass[letterpaper, 10 pt, journal, twoside]{IEEEtran} 

\IEEEoverridecommandlockouts                              

\usepackage[
    maxnames=3,
    minnames=3,
    style=ieee]{biblatex}
\AtEveryBibitem{%
  \clearlist{publisher} 
  \clearfield{issn} 
  \clearfield{doi} 
  \clearfield{isbn} 
  \clearfield{pages} 
  \clearfield{volume} 
  \clearfield{month} 
  \clearfield{number} 

  \ifentrytype{online}{}{
    \clearfield{url}
  }
}

\DeclareSourcemap{ 
    \maps[datatype=bibtex]{
      \map{
           \step[fieldsource=journal, match=Robotics: Science and Systems, replace=\regexp{RSS}]
           \step[fieldsource=booktitle, match=Robotics: Science and Systems, replace=\regexp{RSS}]
           \step[fieldsource=journal, match=International Conference on Robotics and Automation, replace=\regexp{ICRA}]
           \step[fieldsource=booktitle, match=International Conference on Robotics and Automation, replace=\regexp{ICRA}]
           \step[fieldsource=journal, match=International Conference on Learning Representations, replace=\regexp{ICLR}]
           \step[fieldsource=booktitle, match=International Conference on Learning Representations, replace=\regexp{ICLR}]
           \step[fieldsource=journal, match=\regexp{International}, replace=\regexp{Int.}]
           \step[fieldsource=booktitle, match=\regexp{International}, replace=\regexp{Int.}]
           \step[fieldsource=journal, match=\regexp{Conference}, replace=\regexp{Conf.}]
           \step[fieldsource=booktitle, match=\regexp{Conference}, replace=\regexp{Conf.}]
           \step[fieldsource=journal, match=\regexp{Transactions}, replace=\regexp{Trans.}]
           \step[fieldsource=booktitle, match=\regexp{Transactions}, replace=\regexp{Trans.}]
           \step[fieldsource=booktitle, match=Robotics and Automation Letters, replace=\regexp{RA-L}]
           \step[fieldsource=journal, match=Robotics and Automation Letters, replace=\regexp{RA-L}]
           \step[fieldsource=booktitle, match=Advances in Neural Information Processing Systems, replace=\regexp{NeurIPS}]
           \step[fieldsource=journal, match=Advances in Neural Information Processing Systems, replace=\regexp{NeurIPS}]
           \step[fieldsource=booktitle, match=Neural Information Processing Systems, replace=\regexp{NeurIPS}]
           \step[fieldsource=journal, match=Neural Information Processing Systems, replace=\regexp{NeurIPS}]
          }
    }      
}

\AtBeginBibliography{\small}
\usepackage{hyperref}
\usepackage{subcaption}
\usepackage{graphicx} 
\usepackage{times} 
\usepackage{amsmath} 
\usepackage{amssymb}  
\usepackage{blindtext}
\usepackage[table,xcdraw]{xcolor}
\usepackage{booktabs}
\usepackage[capitalize]{cleveref}
\usepackage{algorithm}
\usepackage{algpseudocode}
\usepackage{siunitx}
\usepackage{multirow}
\usepackage{dsfont}
\usepackage[frozencache=true,cachedir=minted-cache]{minted} 

\makeatletter
\let\FV@ListProcessLineOrig\FV@ListProcessLine
\def\FV@ListProcessLine#1{%
  \ifx\FV@Line\empty
    \hbox{}\vspace{-6pt}%
  \else
    \FV@ListProcessLineOrig{#1}%
  \fi}
\makeatother

\usepackage{comment}

\usepackage{pgfplots}
\DeclareUnicodeCharacter{2212}{−}
\usepgfplotslibrary{groupplots,dateplot}
\usepackage{tikz,tikz-3dplot}
\usetikzlibrary{patterns,shapes.arrows}
\pgfplotsset{compat=newest}

\newcommand{\algname}{\mbox{RTN-MPC}}
\newcommand{\emphalgname}{\emph{\algname}}

\newcommand{\rom}[1]{\uppercase\expandafter{\romannumeral #1\relax}}

\newcommand{\wfr}[0]{\ensuremath{W}} 
\newcommand{\bfr}[0]{\ensuremath{B}} 

\newcommand{\bm}[1]{\boldsymbol{#1}}
\newcommand{\mat}[1]{\begin{bmatrix}#1\end{bmatrix}}

\definecolor{cred}{HTML}{E31A1C}
\definecolor{cgreen}{HTML}{33A02C}
\definecolor{cblue}{HTML}{1F78B4}
\definecolor{cgrey}{HTML}{5D6262}
\definecolor{clightgrey}{HTML}{E4E3E1}
\definecolor{cpink}{HTML}{FB9A99}

\usepackage{xspace}

\renewcommand{\baselinestretch}{0.945}
\makeatletter
\DeclareRobustCommand\onedot{\futurelet\@let@token\@onedot}
\def\@onedot{\ifx\@let@token.\else.\null\fi\xspace}

\def\etal{\emph{et al}\onedot}
\makeatother

\newcommand\changes[1]{#1} 

\title{
Real-time Neural MPC:\\Deep Learning Model Predictive Control for Quadrotors and Agile Robotic Platforms
}

\author{Tim Salzmann$^{\scriptstyle 1}$, Elia Kaufmann$^{\scriptstyle 2}$, Jon Arrizabalaga$^{\scriptstyle 1}$, Marco Pavone$^{\scriptstyle 3}$, Davide Scaramuzza$^{\scriptstyle 2}$ and Markus Ryll$^{\scriptstyle 1,5}$
\thanks{Manuscript received: September, 2nd, 2022; Revised November, 27th, 2022; Accepted January, 24th, 2023.}
\thanks{This paper was recommended for publication by Editor Pauline Pounds upon evaluation of the Associate Editor and Reviewers' comments.} 
\thanks{$^{1}$Tim Salzmann, Jon Arrizabalaga and Markus Ryll are with the Technical University of Munich
        {\tt\scriptsize \{Tim.Salzmann, Jon.Arrizabalaga, Markus.Ryll\}@tum.de}}%
\thanks{$^{2}$Elia Kaufmann and Davide Scaramuzza are with the University of Zurich
        {\tt\scriptsize \{ekaufmann, sdavide\}@ifi.uzh.ch}}%
\thanks{$^{3}$Marco Pavone is with the Stanford University and NVIDIA Research
        {\tt\scriptsize pavone@stanford.edu}}%
\thanks{$^{5}$Munich Institute of Robotics and Machine Intelligence (MIRMI)}%
\thanks{Digital Object Identifier (DOI): 10.1109/LRA.2023.3246839}
}

\begin{document}

\markboth{IEEE Robotics and Automation Letters. Preprint Version. Accepted January, 2023}
{Salzmann \MakeLowercase{\textit{et al.}}: Real-time Neural MPC} 

\maketitle

\begin{abstract}
Model Predictive Control (MPC) has become a popular framework in embedded control for high-performance autonomous systems. 
However, to achieve good control performance using MPC, an accurate dynamics model is key. 
To maintain real-time operation, the dynamics models used on embedded systems have been limited to simple first-principle models, which substantially limits their representative power. 
In contrast to such simple models, machine learning approaches, specifically neural networks, have been shown to accurately model even complex dynamic effects, but their large computational complexity hindered combination with fast real-time iteration loops. 
With this work, we present \textit{Real-time Neural MPC}, a framework to efficiently integrate large, complex neural network architectures as dynamics models within a model-predictive control pipeline.
Our experiments, performed in simulation and the real world onboard a highly agile quadrotor platform, demonstrate the capabilities of the described system to run learned models with, previously infeasible, large modeling capacity using gradient-based online optimization MPC. Compared to prior implementations of neural networks in online optimization MPC we can leverage models of over 4000 times larger parametric capacity in a 50Hz real-time window on an embedded platform.
Further, we show the feasibility of our framework on real-world problems by reducing the positional tracking error by up to 82\% when compared to state-of-the-art MPC approaches without neural network dynamics.\\

\vspace{-0.6em}
\noindent Framework Code: \hspace{0.8em}\url{https://github.com/TUM-AAS/ml-casadi}\\
Experimental Code: \url{https://github.com/TUM-AAS/neural-mpc}

\end{abstract}


\section{INTRODUCTION}

\IEEEPARstart{M}{odel} Predictive Control (MPC) is one of the most popular frameworks in embedded control thanks to its ability to simultaneously address actuation constraints and performance objectives through optimization. 
Due to its predictive nature, the performance of MPC hinges on the availability of an accurate dynamics model of the underlying system. 
This requirement is exacerbated by strict real-time constraints, effectively limiting the choice of dynamics models on embedded platforms to simple first-principle models. 
Combining MPC with a more versatile and efficient dynamics model would allow for an  
improvement in performance, safety and operation closer to the robot's physical limits.

Precise dynamics modeling of autonomous systems is challenging, e.g. when the platform approaches high speeds and accelerations or when in contact with the environment. 
Accurate modeling is especially challenging for autonomous aerial systems, as 
high speeds and accelerations can lead to complex aerodynamic effects~\cite{Bauersfeld2021NeuroBEM:Model},
and operating in close proximity to obstacles with an aerial vehicle requires modeling of interaction forces, e.g. ground effect.
Data-driven approaches, in particular neural networks, demonstrated the capability to accurately model highly nonlinear dynamical effects \cite{Bauersfeld2021NeuroBEM:Model, Saviolo2022Physics-InspiredTracking}. 
However, due to their large computational complexity, the integration of such models into embedded MPC pipelines remains challenging due to high frequency real-time requirements.
To overcome this problem prior works have relied on one of two strategies: 

\begin{figure}
    \centering
    \includegraphics[width=\linewidth]{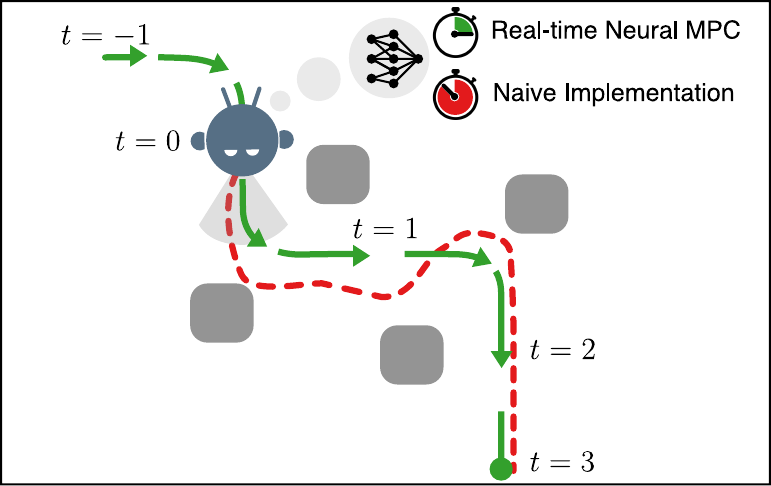}
    \caption{Embedded Model Predictive Control using a neural network as learned dynamics model. Naive integration of the neural network in the MPC optimization loop would lead to extensive optimization times (red) resulting in instabilities. Our approach can handle complex larger learning models while being real-time capable (green).}\vspace{-0.4em}
    \label{fig:hero_figure}
\end{figure}

\hspace{-0.4em}(\rom{1})~Largely reducing the model's capacity to the point where a lot of the predictive performance is lost but real-time speeds can be achieved~\cite{Saviolo2022Physics-InspiredTracking, Torrente2021Data-DrivenQuadrotors, Kabzan2019Learning-BasedRacing, Chee2022KNODE-MPC:Robots, Spielberg2021NeuralFriction}. Commonly, the model is reduced to a Gaussian Process (GP) with few supporting points~\cite{Torrente2021Data-DrivenQuadrotors, Kabzan2019Learning-BasedRacing} or small neural networks~\cite{Chee2022KNODE-MPC:Robots, Williams2017InformationLearning, Spielberg2021NeuralFriction}. Still, these methods are exclusively applied off-device on a powerful CPU.

\hspace{-0.4em}(\rom{2})~A control strategy different to online optimized MPC is used which are either non-predictive \cite{Shi2019NeuralDynamics, faessler2017differential}, do not use online optimization \cite{Chua2018DeepModels, Lambert2019Low-LevelLearning, Williams2017InformationLearning}, or learn the controller end-to-end \cite{Maddalena2020AData, Nubert2020SafeControl, Wang2022ModelConverters, Winqvist2020OnControl, kaufmann2020RSS, Henaff2019Model-PredictiveTraffic}.


In this paper, we present an efficient framework, \textit{Real-time Neural MPC} (\emphalgname{}), that allows for the integration of large-capacity, data-driven dynamics models in online optimization Model Predictive Control and its deployment in real-time on embedded devices.
Specifically, the framework enables the integration of arbitrary neural network architectures as dynamics constraints into the MPC formulation.
To this end, \emphalgname{} leverages CPU or GPU parallelized local approximations of the data-driven model.
Compared to a naive integration of a deep network into an MPC framework, our approach allows unconstrained model architecture selection, embedded real-time capability for larger models, and GPU acceleration, without a decrease in performance.

\subsection*{Contribution}
Our contribution is threefold:
First, we formulate the computational paradigm for \emphalgname{}, an MPC framework which uses deep learning models in the prediction step. By separating the computationally heavy data-driven model from the MPC optimization we can leverage efficient online approximations which allow for larger, more complex models while retaining real-time capability.
Second, we compare and ablate the MPC problem with and without our \emphalgname{} paradigm demonstrating improved real-time capability on CPU, which is further enhanced when GPU processing is available.
Finally, we evaluate our approach on multiple simulation-based and real-world experiments using a high speed quadrotor in aggressive and close-to-obstacle maneuvers. All while running large models, multiple magnitudes higher in capacity compared to state-of-the-art algorithms, in a real-time window.

To the best of the authors' knowledge, this is the first approach enabling data-driven models, in a real-time \textbf{on-board} gradient-based MPC setting on agile platforms. Further, it scales to large models, vastly extending simple two or three-layer networks, on an off-board CPU or GPU enabling model sizes deemed unfit for closed-loop MPC \cite{Bauersfeld2021NeuroBEM:Model, Saviolo2022Physics-InspiredTracking}.
The introduced framework, while demonstrated on agile quadrotors, can be applied broadly and benefit any controlled agile system such as autonomous vehicles or robotic arms


\vspace{1em}
\section{Related Work}\label{sec:related_work}
With the advent of deep learning, there has been a considerable amount of research that aims to combine the representational capacity of deep neural networks with system modeling and control. 
In the following, we provide a brief overview of prior work that focuses on learning-based dynamics modeling, and data-driven control.

{\bf Data-driven Dynamics Models.} 
Thanks to their ability to identify patterns in large amounts of data, deep neural networks represent a promising approach to model complex dynamics. 
Previous works that leverage the representational power of deep networks for such modeling tasks include aerodynamics modeling of quadrotors~\cite{Bauersfeld2021NeuroBEM:Model, Saviolo2022Physics-InspiredTracking, Bansal2016LearningControl} and helicopters~\cite{Punjani2015DeepModels}, turbulence prediction~\cite{li2020fourier}, tire friction modeling~\cite{Spielberg2019NeuralDriving}, and actuator modeling~\cite{hwangbo2019learning}. 
Although these works demonstrated that neural networks can learn system models that are able to learn the peculiarities of real-world robotic systems, they were restricted to simulation-only use cases or employed the network predictions as simple feedforward components in a traditional control pipeline: Saviolo \etal~\cite{Saviolo2022Physics-InspiredTracking} had to revert their accurate physics-inspired model to a simple multi-layer-perceptron for closed-loop control.

{\bf Data-Driven Control.} 
\begin{table}
\centering
\fontsize{6}{6}\selectfont
\renewcommand\arraystretch{1.2}
\renewcommand\tabcolsep{5pt}
\caption{Comparison of state-of-the-art data-driven MPC algorithms and their modeling capacity used for real-time (RT) applications. Prior works use models with small modeling capacity on high-end CPUs while our approach can leverage powerful models on an embedded platform.}
\begin{tabular}{@{}l|cccc@{}}
\toprule
~ & Model Architecture & RT Complexity & Parameters & RT Platform \\ \midrule
DD-MPC~\cite{Torrente2021Data-DrivenQuadrotors} & Gaussian Process & 20 Sup. Points & 120 & Intel i7 \\
NNMPC~\cite{Spielberg2021NeuralFriction} &  MLP & 2 Layer (64x64) & 4096 & Intel i7 \\
KNODE-MPC~\cite{Chee2022KNODE-MPC:Robots} & MLP & 1 Layer (32) & 32 & Intel i7 \\
PI-TCN~\cite{Saviolo2022Physics-InspiredTracking} & MLP & 3 Layer(64x32x32) & 3072 & Laptop \\ \midrule
Ours (\emphalgname) & Diverse & Diverse & up to 11M & Jetson ARM/GPU \\
\bottomrule
\end{tabular}\label{tab:sota}
\vspace{1em}
\end{table}
Leveraging the power of learned models in embedded control frameworks has been extensively researched in recent years. 
Most approaches have focused on combining the learned model with a simple reactive control scheme, such as the ``Neural Lander'' approach~\cite{Shi2019NeuralDynamics}. 
Neural Lander uses a learned model of the aerodynamic ground-effect to substantially improve a set-point controller in near-hover conditions. 
In~\cite{Lenz2015DeepMPC:Control}, a learned recurrent dynamics model formulates a model-based control problem.  
While this approach allowed the system to adapt online to changing operating conditions, it cannot account for system constraints such as limited actuation input. 
Recent approaches that integrate the modeling strengths of data-driven approaches in the MPC framework propose the use of Gaussian Processes (GP) as a learned residual model for race cars~\cite{Kabzan2019Learning-BasedRacing} and quadrotors~\cite{Torrente2021Data-DrivenQuadrotors}.
For Gaussian Processes, both their complexity and accuracy scale with the number of inducing points and with their dimensionality, limiting their performance on embedded systems.
The approaches of Chee \etal~\cite{Chee2022KNODE-MPC:Robots},  Williams \etal \cite{Williams2017InformationLearning} and Spielberg \etal~\cite{Spielberg2021NeuralFriction} follow the approach of \cite{Torrente2021Data-DrivenQuadrotors} but model the quadrotor's dynamics residual using a small neural network for different applications.
In \cref{tab:sota}, we compare existing data-driven MPC approaches based on their modeling capacity. All state-of-the-art models are severely limited by the small modeling capacity of either GPs with a small number of supporting points or small two- or three-layer neural networks.

With the rise of deep reinforcement learning (RL), a new class of controllers for robotic systems has emerged that directly maps sensory observations to actions. 
Popular instances of such RL controllers are imitation learning of an expert controller~\cite{Maddalena2020AData, Nubert2020SafeControl, kaufmann2020RSS} as well as Model-free and Model-based reinforcement learning~\cite{andrychowicz2020learning, Lambert2019Low-LevelLearning, Williams2017InformationLearning, Henaff2019Model-PredictiveTraffic, Chua2018DeepModels}.
Although such approaches achieve high control frequencies and may outperform online MPC approaches, they commonly require training in simulation, do not allow for tuning without costly retraining, and often discard the optimality, robustness and generalizability of an online optimized MPC framework.


Our work is inspired by~\cite{Kabzan2019Learning-BasedRacing, Torrente2021Data-DrivenQuadrotors, Chee2022KNODE-MPC:Robots, Williams2017InformationLearning, Spielberg2021NeuralFriction, Saviolo2022Physics-InspiredTracking} but replaces the Gaussian Process dynamics of~\cite{Kabzan2019Learning-BasedRacing, Torrente2021Data-DrivenQuadrotors} or the small neural networks of~\cite{Chee2022KNODE-MPC:Robots, Spielberg2021NeuralFriction} with networks of higher modeling capacity \cite{Bauersfeld2021NeuroBEM:Model, Saviolo2022Physics-InspiredTracking} and uses gradient-based optimization as opposed to a sampling-based scheme~\cite{Williams2017InformationLearning}.
The resulting framework allows a combination of the versatile modeling capabilities of deep neural networks with state-of-the-art embedded optimization software without tightly constraining the choice of network architecture.

\section{Problem Setup}\label{sec:npc}
In its most general form, MPC solves an optimal control problem (OCP) by finding an input command $\mathbf{u}$ which minimizes a cost function $\mathcal{L}$ subject to its system dynamics model $\dot{\mathbf{x}} = f(\mathbf{x}, \mathbf{u})$ while accounting for constraints on input and state variables for current and future timesteps.
Traditionally, the model $f$ is manually derived from first principles using ``simple'' differential-algebraic equations (DAE) which often neglect complicated dynamics effects such as aerodynamics or friction as they are hard or computationally expensive to formalize. 
Following prior works \cite{Torrente2021Data-DrivenQuadrotors, Chee2022KNODE-MPC:Robots, Kabzan2019Learning-BasedRacing, Saviolo2022Physics-InspiredTracking}, we partition $f$ into a mathematical combination of first principle DAEs $f_\mathcal{F}$ and a learned data-driven model $f_\mathcal{D}$. 
This enables more general models extending the capability of DAE dynamics models.
To solve the aforementioned OCP, we approximate it by discretizing the system into $N$ steps of step size $\delta t$ over a time horizon~$T$ using direct multiple shooting \cite{Bock1984AProblems} which leads to the following nonlinear programming (NLP) problem
\begin{equation}\label{eq:nlp}
\begin{split}
    \min_{\mathbf{u}} &\sum^{N-1}_{k=0} \mathcal{L}(\mathbf{x_k}, \mathbf{u_k}) \\
    \text{subject to} \qquad &\mathbf{x}_{k=0} = \mathbf{x}_{0} \\
    &\mathbf{x}_{k+1} = \phi(\mathbf{x}_k, \mathbf{u}_k, f, \delta t) \\
    & f(\mathbf{x}_k, \mathbf{u}_k) = f_{\mathcal{F}}(\mathbf{x}_k, \mathbf{u}_k) + f_{\mathcal{D}}(\mathbf{x}_k, \mathbf{u}_k) \\
    &g(\mathbf{x}_k, \mathbf{u}_k) \leq 0
\end{split}
\end{equation}
where $\mathbf{x}_{0}$ denotes the initial condition and $g$ can incorporate (in-)equality constraints, such as bounds in state and input variables. $\phi$ is the numerical integration routine to discretize the dynamics equation where commonly a 4th order \textit{Runge-Kutta} algorithm is used involving $E=4$ evaluations of the dynamics function $f$.
To leverage advancements in embedded solvers, the NLP is optimized using sequential quadratic programming (SQP) with $\mathbf{\omega}$ being the SQP iterate $\mathbf{\omega}^i = [\mathbf{x}^i_0, \mathbf{u}^i_0, \ldots,  \mathbf{x}^i_{N-1}, \mathbf{u}^i_{N-1}]$.

\section{Bringing Neural MPC to Onboard Real-time}
In this section, we lay down the key concepts to speed up the optimization times of MPC control with neural networks. The key insight in \cref{sec:approx} is that local approximations of the learned dynamics are sufficient to keep alike performance while drastically improving the generation process of the optimization problem. This insight is utilized in a three-phased embedded real-time optimization procedure in \cref{sec:rtn-mpc}.
\subsection{Locally Approximated Continuity Quadratic Program}\label{sec:approx}
Due to advances in embedded optimization solvers, SQP has become a well-suited framework to efficiently solve NLPs resulting from multiple shooting approximations of OCPs.
This involves repetitively approximating and solving \cref{eq:nlp} as a quadratic program (QP). The solution to the QP leads to an update on the iterate $\mathbf{\omega}^{i+1} = \mathbf{\omega}^{i} + \Delta \mathbf{\omega}^i$ where the step $\Delta \mathbf{\omega}^i$ is given by solving the following QP
\begin{align*}
	\min_{\Delta \mathbf{\omega}^i}&
	\sum_{k=0}^{N-1} \begin{bmatrix}\mathbf{q}_k \\ \mathbf{r}_k\end{bmatrix}^\top \begin{bmatrix}\Delta \mathbf{x}_k \\ \Delta \mathbf{u}_k \end{bmatrix} + \begin{bmatrix}\Delta \mathbf{x}_k \\ \Delta \mathbf{u}_k \end{bmatrix}^\top \mathbf{H}_k \begin{bmatrix}\Delta \mathbf{x}_k \\ \Delta \mathbf{u}_k\end{bmatrix} \\
    \text{subject to} & \\
    \Delta \mathbf{x}_{k+1} &= \mathbf{A}_k \Delta \mathbf{x}_k + \mathbf{B}_k \Delta \mathbf{u}_k + \bar{\mathbf{\phi}}_k - \mathbf{x}_{k+1}\;, \addtocounter{equation}{1}\tag{\theequation} \label{eq:continuity}\\
    &\hspace{9.7em} \quad k=0,\ldots,N-1\; , \\
    -\bar{\mathbf{g}}_k &\geq G_k^x \Delta \mathbf{x}_k + G^u_k \Delta \mathbf{u}_k\;, \quad k=0,\ldots,N\; , \addtocounter{equation}{1}\tag{\theequation} \label{eq:consistency}
\end{align*}
where $\mathbf{q}_k = \frac{\delta}{\delta \mathbf{x}^i_k}\mathcal{L}(\mathbf{x}^i_k, \mathbf{u}^i_k)$, $\mathbf{r}_k = \frac{\delta}{\delta \mathbf{u}^i_k}\mathcal{L}(\mathbf{x}^i_k, \mathbf{u}^i_k)$ linearize the cost function and, under given circumstances, the hessian $\mathbf{H}_k$ can be approximated by the \textit{Gauss-Newton} algorithm. $\bar{\mathbf{\phi}}_k$ and $\bar{\mathbf{g}}_k$ are shorthand notations for the function evaluations $\phi(\mathbf{x}_k^i, \mathbf{u}_k^i, f, \delta t)$ and $g(\mathbf{x}_k^i, \mathbf{u}_k^i)$.
The main computational burden lies in the parameter computation of the continuity condition \cref{eq:continuity}. Specifically for each shooting node $k=0,\ldots,N-1$ we need to compute
\begin{equation*}
\begin{split}
	\mathbf{A}_k = \frac{\delta}{\delta \mathbf{x}_k^i} \phi(\mathbf{x}_k^i, \mathbf{u}_k^i, f, &\delta t)\; ,  
    \quad\mathbf{B}_k = \frac{\delta}{\delta \mathbf{u}_k^i} \phi(\mathbf{x}_k^i, \mathbf{u}_k^i, f, \delta t)\; , \\
    \bar{\mathbf{\phi}}_k &= \phi(\mathbf{x}_k^i, \mathbf{u}_k^i, f, \delta t)\; .
\end{split}
\end{equation*}
Leading to $N * E * 2$ evaluations of the partial differentiations 
\begin{equation*}
\delta f(\mathbf{x}, \mathbf{u}) = \delta f_\mathcal{N}(\mathbf{x}, \mathbf{u}) + \delta f_\mathcal{D}(\mathbf{x}, \mathbf{u})
\end{equation*}
and $N*E$ function evaluations 
\begin{equation*}
f(\mathbf{x}, \mathbf{u}) = f_\mathcal{N}(\mathbf{x}, \mathbf{u}) + f_\mathcal{D}(\mathbf{x}, \mathbf{u})
\end{equation*}
of the dynamics equation. For computational heavy data-driven dynamics models $f_\mathcal{D}$ this leads to extensive processing times generating the QP.

The learned data-driven dynamics $f_{\mathcal{D}}$ are assumed to be accurate over the entire input space of states and controls present in the training dataset. However, to create the QP continuity condition we only require the model and its differentiations to be accurate in and around specific input values $\mathbf{\omega}^i$.
Thus, to speed up the QP generation we replace the computationally heavy globally valid data-driven dynamics equation $f_{\mathcal{D}}$ with a computationally light locally valid approximation up to second order around the current iterate
\begin{equation}\label{eq:approx}
\begin{split}
    f_{\mathcal{D}}^*(\mathbf{x}, \mathbf{u}) &\approx \bar{\mathbf{f}}_{\mathcal{D}}^i + \mathbf{J}_{\mathcal{D}, k}^i \begin{bmatrix}\mathbf{x} - \mathbf{x}_k^i \\ \mathbf{u} - \mathbf{u}_k^i \end{bmatrix} \\ &+ \frac{1}{2} \begin{bmatrix}\mathbf{x} - \mathbf{x}_k^i \\ \mathbf{u} - \mathbf{u}_k^i \end{bmatrix}^\top \mathbf{H}_{\mathcal{D}, k}^i \begin{bmatrix}\mathbf{x} - \mathbf{x}_k^i \\ \mathbf{u} - \mathbf{u}_k^i \end{bmatrix}\; .
\end{split}
\end{equation}

The required differentiations are readily available as submatrices of $\mathbf{J}_{\mathcal{D}, k}^i$ for first-order approximations or as submatrices of a Tensor multiplication and sum for second-order approximations. The induced error of this computational simplification is of second order for a first-order approximation and of third order for a second-order approximation in the size of state and control changes between nodes. We will experimentally demonstrate this error to be neglectable for agile platforms where $\delta t$ is small in \cref{sec:experiments}.

Applying \cref{eq:approx}, the QP creation becomes independent of the complexity and architecture of the data-driven dynamics model. Further, with $\mathbf{J}_{\mathcal{D}, k}^i$ and $\mathbf{H}_{\mathcal{D}, k}^i$ being the single interfaces between the SQP optimization and the data-driven dynamics model, we are free to optimize the approximation process independent of the NLP framework; passing them as parameters to the continuity condition procedure of the QP generation. 
As $f_{\mathcal{D}}$ is a neural network model commonly consisting of large matrix multiplications we are therefore free to use algorithms and hardware optimized for neural network evaluation and differentiation. Those capabilities are readily available in modern machine learning tools such as PyTorch \cite{Paszke2019PyTorch:Library} and TensorFlow \cite{Abadi2016TensorFlow:Systems}. This enables us to calculate the Jacobians and Hessians for all shooting nodes $N$ as a single parallelized batch on CPU or GPU.

\subsection{Real-time Neural MPC}\label{sec:rtn-mpc}
\begin{figure}
    \centering
    \includegraphics[width=\linewidth]{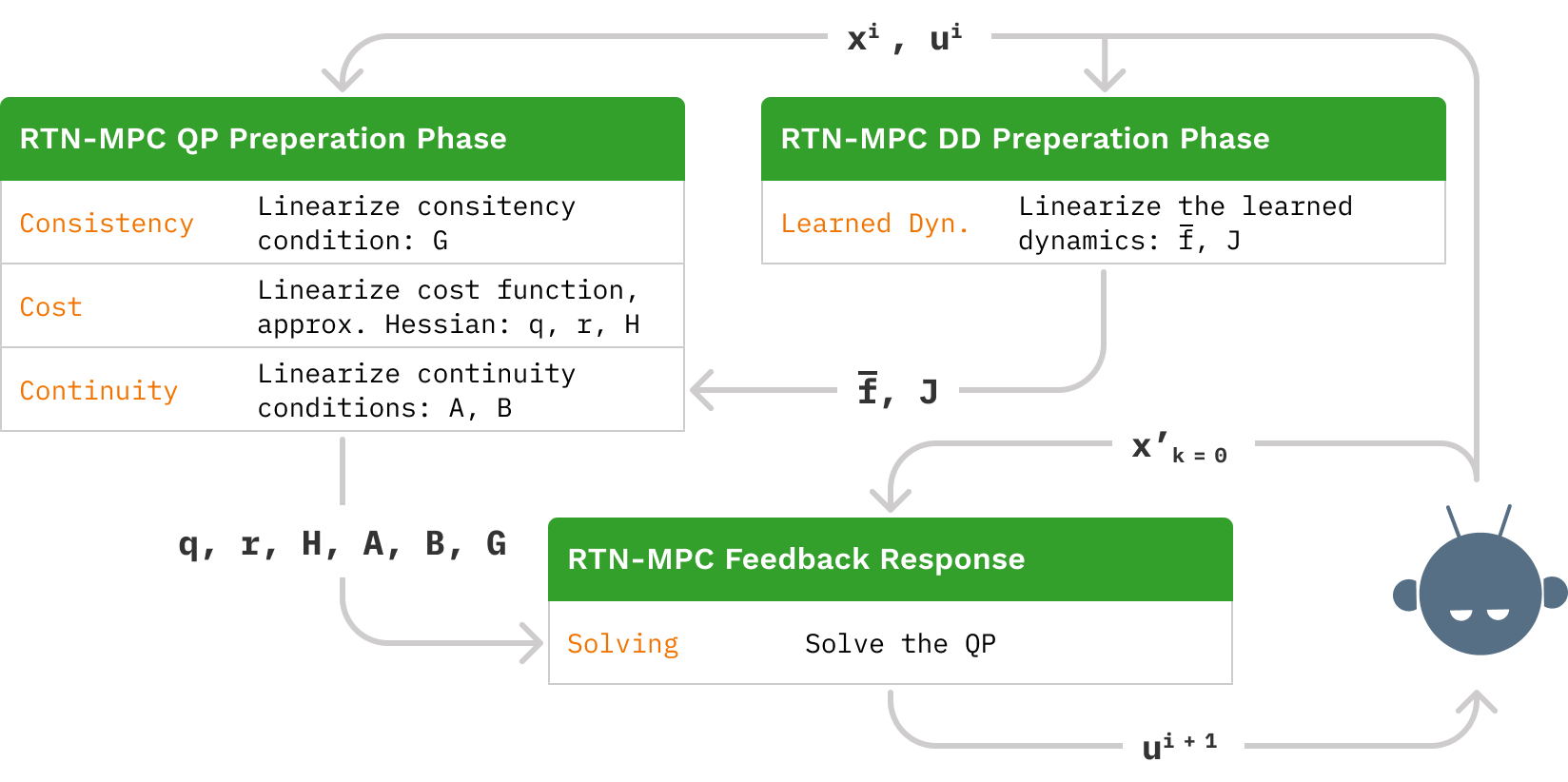}
    \caption{Data flow for our \emphalgname{} algorithm. The data-driven (DD) preparation phase is performed efficiently using optimized machine learning batch-differentiation tools on CPU or GPU.}
    \label{fig:rti_neural_mpc}
\end{figure}

Even without a data-driven dynamics model, solving the SQP until convergence is computationally too costly in real-time for agile robotic platforms. To account for this shortcoming, MPC applications subjected to fast dynamics are commonly solved using a real-time-iteration scheme~(RTI)~\cite{Diehl2002Real-timeEquations}, where only a single SQP iteration is executed - one quadratic problem is constructed and solved as a potentially sub-optimal but timely input command is preferred over an optimal late one. 
As shown in \cref{fig:rti_neural_mpc}, \emphalgname{} divides the real-time optimization procedure into three parts: 
\textit{QP Preparation Phase}, \textit{Data-Driven Dynamics Preparation Phase} and \textit{Feedback Response}. 

With available iterate $\omega^i$, the data-driven dynamics preparation phase calculates $\bar{f}_{\mathcal{D}}^i$ and $\mathbf{J}_{\mathcal{D}, k}^i$ using efficient batched differentiates of the data-driven dynamics on CPU or GPU.

Meanwhile, the QP preparation phase constructs a QP by linearizing around $\mathbf{x}^i$ and control $\mathbf{u}^i$ using a first-order approximation $f_{\mathcal{D}}^*(\mathbf{x}, \mathbf{u})$ for the continuity condition parametrized by the result of the data-driven dynamics preparation phase.

Once a new disturbed state $\mathbf{x}'_{k=0}$ is sensed, the feedback response phase solves the pre-constructed QP using the disturbed state as input. The iterate $\omega$ is adjusted with the QP result and the optimized command $\mathbf{u}$ is sent to the actuators.

\subsection{Implementation}
To demonstrate the applicability of the \emphalgname{} paradigm, we provide a implementation using CasADi~\cite{Andersson2019CasADi:Control} and acados \cite{Verschueren2021Acados:Control} as the optimization framework and PyTorch~\cite{Paszke2019PyTorch:Library} as ML framework. 
This enables the research community to use arbitrary neural network models, trainable in PyTorch and usable in CasADi.

Further, we will compare our \emphalgname{} approach against a naive implementation of a neural network data-driven MPC as applied in~\cite{Spielberg2021NeuralFriction, Chee2022KNODE-MPC:Robots, Saviolo2022Physics-InspiredTracking}. Here, the learned model is directly constructed in CasADi in the form of trained weight matrices and activation functions. Subsequently, the QP generation and automatic differentiation engine in CasADi has to deal with the full neural-network structure for which it is lacking optimized algorithms while being confined to the CPU.

\section{Runtime Analysis}\label{sec:experiments_runtime}
We demonstrate the computational advantage of our proposed \emphalgname{} paradigm compared to a naive implementation of a data-driven dynamics model in online MPC. Thus, we construct an experimental problem in which the nominal dynamics is trivial while the data-driven dynamics can be arbitrarily scaled in computational complexity.
As such the nominal dynamics model is a double integrator on a position $p$ while the data-driven dynamics is a neural network of variable architecture. To solely focus on the computational complexity of the data-driven dynamics, rather than modeling accuracy, the networks are not trained but weights are manually adjusted to force a zero output.
\begin{align}
\begin{split}
\dot{\bm{x}} = 
\begin{bmatrix}
\dot{p} \\  
\ddot{p} \\
\end{bmatrix} &= 
f_{\mathcal{F}}(\bm{x}, u) = 
\begin{bmatrix}
\dot{p} \\  
u \\
\end{bmatrix} \; , \\
f(\bm{x}, u) &= f_{\mathcal{F}}(\bm{x}, u) + \overbrace{f_{\mathcal{D}}(\bm{x}, u)}^{0}\; . \\
\end{split}
\end{align}
We use an explicit \textit{Runge-Kutta} method of 4th order $\phi(\bm{x}, u, f, \delta t) = RK4(\bm{x}, u, f, \delta t)$ to numerically integrate $f$.

In this experiment, we simulate the system without any model-plant-mismatch to focus solely on runtime.
The optimization problem is solved by constructing the multiple shooting scheme with $N=10$ nodes.

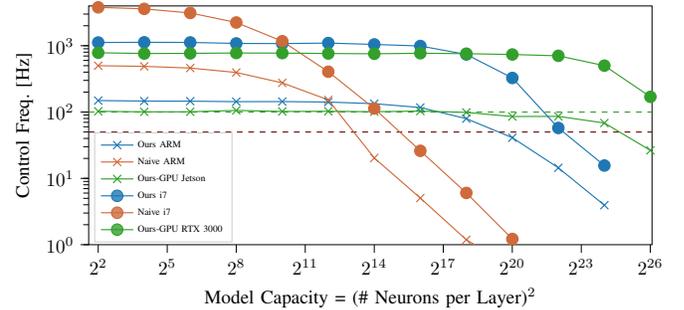
\begin{figure}
    \centering
    \resizebox{\linewidth}{!}{
\begin{tikzpicture}

\definecolor{color0}{rgb}{0.2,0.627450980392157,0.172549019607843}
\definecolor{color1}{rgb}{0.43921568627451,0.0431372549019608,0.0470588235294118}
\definecolor{color2}{rgb}{0.12156862745098,0.470588235294118,0.705882352941177}
\definecolor{color3}{rgb}{0.811764705882353,0.415686274509804,0.231372549019608}

\begin{axis}[
height=6cm,
legend cell align={left},
legend style={
  font=\tiny,
  fill opacity=0.8,
  draw opacity=1,
  text opacity=1,
  at={(0.01,0.01)},
  anchor=south west,
  draw=white!80!black
},
log basis x={2},
log basis y={10},
tick align=outside,
tick pos=left,
width=12cm,
x grid style={white!69.0196078431373!black},
xlabel={Model Capacity = (\# Neurons per Layer)${}^2$},,
xmin=3, xmax=70464307,
xmode=log,
xtick style={color=black},
y grid style={white!69.0196078431373!black},
ylabel={Control Freq. [Hz]},
ymin=1, ymax=3977.94725789993,
ymode=log,
ytick style={color=black}
]
\addplot [semithick, color0, dashed, forget plot]
table {%
3 100
70464307.0000001 100
};
\addplot [semithick, color1, dashed, forget plot]
table {%
3 50
70464307.0000001 50
};
\addplot [semithick, color2, mark=x, mark size=3, mark options={solid}]
table {%
4 149.004903057028
16 146.418717756137
64 145.850933832118
256 143.379180465856
1024 143.841966106513
4096 141.45906030033
16384 133.713531711705
65536 117.028819592364
262144 79.5121379637692
1048576 41.0584271541107
4194304 14.4254515913872
16777216 3.93943607688031
};
\addlegendentry{Ours ARM}
\addplot [semithick, color3, mark=x, mark size=3, mark options={solid}]
table {%
4 499.487627621229
16 488.477870493346
64 460.60589554778
256 395.090271206897
1024 275.340285488936
4096 152.53544337283
16384 20.3090689394351
65536 5.04830029489157
262144 1.18673335271124
1048576 0.466711939926341
};
\addlegendentry{Naive ARM}
\addplot [semithick, color0, mark=x, mark size=3, mark options={solid}]
table {%
4 102.657008437188
16 100.709820714456
64 100.92934656498
256 105.863253267578
1024 102.102012186455
4096 102.683811939294
16384 101.195772774583
65536 103.393988304607
262144 98.7165844221461
1048576 85.603734338168
4194304 86.0492567772575
16777216 68.3569525830423
67108864 26.4317139530039
};
\addlegendentry{Ours-GPU Jetson}
\addplot [semithick, color2, mark=*, mark size=3, mark options={solid}]
table {%
4 1118.13399787377
16 1126.03873840189
64 1121.0693916783
256 1083.2424670519
1024 1076.81742130424
4096 1096.91742515604
16384 1045.70658114142
65536 990.449022964797
262144 739.00015275794
1048576 326.486030384201
4194304 57.5390659746092
16777216 15.6179460891193
};
\addlegendentry{Ours i7}
\addplot [semithick, color3, mark=*, mark size=3, mark options={solid}]
table {%
4 3788.54342237803
16 3592.82241402367
64 3117.75549767501
256 2239.07462560727
1024 1169.45365323223
4096 405.513422642949
16384 112.982541772093
65536 26.0043653415727
262144 6.03510881744932
1048576 1.22034817782288
};
\addlegendentry{Naive i7}
\addplot [semithick, color0, mark=*, mark size=3, mark options={solid}]
table {%
4 783.817234693669
16 762.964730809772
64 767.252369136184
256 776.698079857825
1024 773.780447870973
4096 763.706852506781
16384 756.285195861714
65536 770.486162840046
262144 758.611534038682
1048576 737.770069532561
4194304 704.588852478657
16777216 500.941735645685
67108864 169.713018444496
};
\addlegendentry{Ours-GPU RTX 3000}
\end{axis}

\end{tikzpicture}}
    \caption{Evaluation of real-time capability for different two-layer model parametric capacities. We evaluate on an embedded platform (Nvidia Jetson Xavier NX) and a laptop machine (Intel i7, Nvidia RTX 3000). Parametric model capacity is approximated by the squared number of neurons per layer. The \emphalgname{} framework can run 4000 times larger models in parametric complexity compared to a naive implementation. To make the results comparable, we define a target run-time window of at least 50Hz (dashed red line) and preferably over 100Hz (dashed green line). However, in a real-world scenario the real-time window is specific to the use-case.}
    \label{fig:results_runtime_2_layer}
\end{figure}

\cref{fig:results_runtime_2_layer} compares two-layer networks with increasing neuron count for a naive implementation and our \emphalgname{} framework. On an embedded system, such as the Nvidia Jetson Xavier NX, our approach enables larger models of factor 60 in parametric complexity on CPU and of factor 4000 on GPU while staying within a real-time window above 50Hz. Running on a desktop, which is the current default in data-driven MPC research~\cite{Torrente2021Data-DrivenQuadrotors, Chee2022KNODE-MPC:Robots, Spielberg2019NeuralDriving, Saviolo2022Physics-InspiredTracking}, we can run two-layer models with more than 150 million parameters above 100Hz on a low-end GPU (Nvidia RTX3000).

\begin{table}
\centering
\fontsize{6}{6}\selectfont
\renewcommand\arraystretch{1.2}
\renewcommand\tabcolsep{1.7pt}
\caption{Runtime comparison between naive implementation and \emphalgname{}. Model complexity and parameter count are increasing from left to right. The naive approach becomes computationally costly in runtime above $\sim$10K parametric capacity with control frequencies dropping below 50Hz. Our approach can scale to powerful networks and complex network architectures showing real-time control frequencies of over 100Hz for over 50K parameters on an embedded device and over 13M on a GPU.}
\begin{tabular}{@{}lr|c|cccccccc|c|@{}}
\toprule
  & & \multicolumn{10}{c|}{Model Configuration} \\\midrule
 & & None & \multicolumn{8}{c|}{MLP} & \changes{CNN}  \\\midrule
 \multirow{2}{*}{Architecture} &  Layers & - &  2 & 2 & 5 & 5 & 12 & 12 & 20 & 50 & 18 \\ 
&  Neurons & - &  16 & 128 & 16 & 128 & 32 & 512 & 512 & 512 & -   \\ \midrule
\multicolumn{2}{r|}{Parameter Count} & 0 & 354 & 17K & 1.2K & 67K & 12K & 182K & 500K & 13M & 12M  \\ \midrule \midrule
  & & \multicolumn{10}{c|}{Control Freq. {[}Hz{]}} \\ \midrule
 \multirow{1}{*}{Naive} & ARM-CPU & \multirow{3}{*}{562} & \bf{403} & 20 & \bf{280} & 6 & 66 & 2 & $<$1 & - & - \\
\multirow{2}{*}{\emphalgname{}\hspace{1em}} & ARM-CPU &  & 148 & \bf{135} & 118 & \bf{102} & \bf{85} & \bf{67} & 11 & - & $<$1 \\
& Jetson-GPU &  & 109 & 107 & 88 & 84 & 63 & 61 & \bf{46} & - & 9 \\ \midrule
\multirow{1}{*}{Naive} & i7-CPU & \multirow{3}{*}{4262} & \bf{2228} & 116 & \bf{1139} & 31 & 168 & 11 & $<$1 & $<<1$ & - \\ 
\multirow{2}{*}{\emphalgname{}\hspace{1em}}  & i7-CPU &  & 1096 & \bf{1071} & 885 & \bf{784} & \bf{588} & \bf{507} & 91 & 39 & 4 \\
& RTX3000-GPU &  & 781 & 770 & 586 & 598 & 363 & 363 & \bf{232} & \bf{117} &  \bf{63} \\ \midrule \midrule
\end{tabular}
\label{tab:results_runtime}
\end{table}

We further evaluate the runtime of a broad range of deep learning architectures in \cref{tab:results_runtime}. While the naive approach has better runtime for small networks, our approach dominates for larger and deeper networks enabling running a 12 layer 512 neurons each network above 50Hz on an embedded CPU and above 500Hz on a desktop CPU. To demonstrate that complex network architectures are easily integrated in the MPC loop using \emphalgname{}, we run a full \changes{CNN ResNet model~\cite{ramzan2020deep} with 18 convolutional layers} in the optimization loop above 50Hz when leveraging the GPU capabilities of our framework.

\section{Experimental Setup}
While the \emphalgname{} framework described in \cref{sec:npc} can be applied to a variety of robotic applications, we will use agile quadrotor flight maneuvers to showcase its potential for real-world problems.

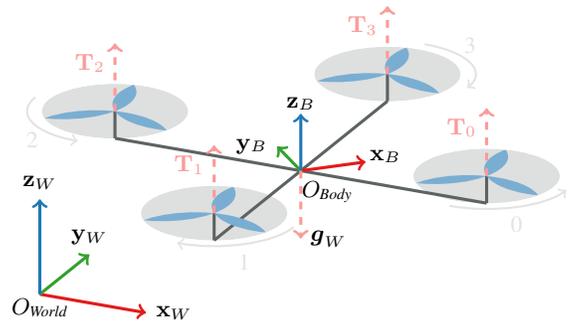
\begin{figure}
    \centering
    \resizebox{\linewidth}{!}{\definecolor{cred}{HTML}{E31A1C}
\definecolor{cgreen}{HTML}{33A02C}
\definecolor{cblue}{HTML}{1F78B4}
\definecolor{cgrey}{HTML}{5D6262}
\definecolor{clightgrey}{HTML}{E4E3E1}
\definecolor{cpink}{HTML}{FB9A99}

\tdplotsetmaincoords{68}{25}

\tikzset
{
  pics/rotor/.style args={#1}{
   code={
    \begin{scope}[tdplot_main_coords,line width=.2pt]
      \def\r{#1}

      \begin{scope}
        \draw [clip, draw=none] (0,0) circle (\r) ;
        \fill [fill=cgrey, opacity=0.2] (0,0) circle (\r) ;
        \draw [draw=none, fill=cblue!60,even odd rule] 
            (40:\r) circle (\r) 
            (160:\r) circle (\r)
            (280:\r) circle (\r) 
            (0,0) circle (2*\r);
      \end{scope}
    \end{scope}
  }}
}

\begin{tikzpicture}[tdplot_main_coords, scale=2]

\draw[very thick, color=cgrey] (-1.4,0,0.8) -- (1.4,0,0.8);
\draw[very thick, color=cgrey] (0,-1.4,0.8) -- (0,1.4,0.8);

\draw[very thick, color=cgrey] (-1.4,0,0.8) -- (-1.4,0,1.0);
\draw[very thick, color=cgrey] (1.4,0,0.8) -- (1.4,0,1.0);
\draw[very thick, color=cgrey] (0,-1.4,0.8) -- (0,-1.4,1.0);
\draw[very thick, color=cgrey] (0,1.4,0.8) -- (0,1.4,1.0);

\pic at (0,1.4,1) {rotor=1.0};
\pic at (1.4,0,1) {rotor=1.0};
\pic at (0,-1.4,1) {rotor=1.0};
\pic at (-1.4,0,1) {rotor=1.0};

\draw[very thick,->,color=cpink, dashed] (0,1.4,1) -- (0,1.4,1.5) node[below left] {\small $\mathbf{T}_3$};
\draw[very thick,->,color=cpink, dashed] (1.4,0,1) -- (1.4,0,1.5) node[below left] {\small $\mathbf{T}_0$};
\draw[very thick,->,color=cpink, dashed] (0,-1.4,1) -- (0,-1.4,1.5) node[below left] {\small $\mathbf{T}_1$};
\draw[very thick,->,color=cpink, dashed] (-1.4,0,1) -- (-1.4,0,1.5) node[below left] {\small $\mathbf{T}_2$};	

\draw[thick,->,color=clightgrey] (0,2.0,1) arc (90:0:0.6) node[midway, above] {\small 3};
\draw[thick,<-,color=clightgrey] (2.0,0,1) arc (0:-90:0.6) node[midway, below] {\small 0};
\draw[thick,->,color=clightgrey] (0.6,-1.4,1) arc (0:-90:0.6) node[midway, below] {\small 1};
\draw[thick,->,color=clightgrey] (-2.0,0,1) arc (180:270:0.6) node[midway, below] {\small 2};

\draw[very thick,->,color=cred,text=black] (0,0,0.8) -- (0.333,0.333,0.8) node[right, shift={(-0.5ex,0.5ex)}]  {\small $\mathbf{x}_{\bfr}$};
\draw[very thick,->,color=cgreen,text=black] (0,0,0.8) -- (-0.333,0.333,0.8) node[above left, shift={(-0.ex,-1.5ex)}] {\small $\mathbf{y}_{\bfr}$};	
\draw[very thick,->,color=cblue,text=black] (0,0,0.8) -- (0,0,1.22) node[above, shift={(0,0, -0.02ex)}] {\small $\mathbf{z}_{\bfr}$};		
\node[draw=none] at (0.2,0,0.68) {\it \small $O_{\text{Body}}$};	

\draw[very thick,->,color=cred,text=black] (-1.5,-1,-0) -- (-0.7,-1,-0) node[right] {\small $\mathbf{x}_{\wfr}$};		
\draw[very thick,->,color=cgreen,text=black] (-1.5,-1,-0) -- (-1.5,-0.2,-0) node[above] {\small $\mathbf{y}_{\wfr}$};		
\draw[very thick,->,color=cblue,text=black] (-1.5,-1,-0) -- (-1.5,-1,0.7) node[above] {\small $\mathbf{z}_{\wfr}$}; 
\node[draw=none] at (-1.5,-1,-0.1) {\it \small $O_{\text{World}}$};



\draw[very thick,->,color=cpink, dashed,text=black] (0,0,0.8) --  (0,0,0.3) node[right] {\small $\bm{g}_{\wfr}$};

\end{tikzpicture}}
    \caption{Quadrotor model with world and body frames and propeller numbering convention. Grey arrows indicate the spinning direction of the individual rotors.}
    \label{fig:quad_model}
\end{figure}

{\bf Notation.}
Scalars are denoted in lowercase $s$, vectors in lowercase bold $\bm{v}$, and matrices in uppercase bold $\bm{M}$.
Coordinate frames such as the World $W$ and Body $B$ frames are defined with orthonormal basis i.e. $\{\bm{x}_W, \bm{y}_W, \bm{z}_W\}$, with the Body frame being located at the center of mass of the quadrotor (see \cref{fig:quad_model}).
A vector from coordinate $\bm{p}_1$ to $\bm{p}_2$ expressed in the $W$ frame is written as $_W\bm{v}_{12}$.
If the vector's origin coincides with the frame it is described in, the frame index is dropped, e.g. the quadrotor position is denoted as $\bm{p}_{WB}$.
Orientations are represented using unit quaternions $\bm{q} = (q_w, q_x, q_y, q_z)$ with $\|\bm{q}\| = 1$, such as the attitude state of the quadrotor body $\bm{q}_{WB}$.
Finally, full SE3 transformations, such as changing the frame of reference from Body to World for a point $\bm{p}_{B1}$, are described by ${}_W\bm{p}_{B1} = {}_W\bm{t}_{WB} + \bm{q}_{WB} \odot \bm{p}_{B1}$.
Note the quaternion-vector product denoted by $\odot$ representing a rotation of the vector~$\bm{v}$ by the quaternion~$\bm{q}$ as in $\bm{q} \odot \bm{v} = \bm{q} \bm{v} \bar{\bm{q}}$, where $\bar{\bm{q}}$ is the quaternion's conjugate.

{\bf Nominal Quadrotor Dynamics Model.}
The nominal dynamics assume the quadrotor to be a 6 degree-of-freedom rigid body of mass $m$ and diagonal moment of inertia matrix $\bm{J}=\mathrm{diag}(J_x, J_y, J_z)$.
Our model is similar to~\cite{Torrente2021Data-DrivenQuadrotors, Falanga2018PAMPC:Quadrotors, Kamel2017ModelSystem} as we write the nominal dynamics $\dot{\bm{x}}$ up to second order derivatives, leaving the quadrotors individual rotor thrusts $T_i\ \forall\ i \in (0, 3)$ as control inputs $\bm{u} \in \mathbb{R}^4$. 
The state space is thus 13-dimensional and its dynamics can be written as:
\begin{align}\label{eq:nom_dyn}
\small
\dot{\bm{x}} =
\begin{bmatrix}
\dot{\bm{p}}_{WB} \\  
\dot{\bm{q}}_{WB} \\
\dot{\bm{v}}_{WB} \\
\dot{\boldsymbol\omega}_B
\end{bmatrix} = 
f_{\mathcal{F}}(\bm{x}, \bm{u}) =
\begin{bmatrix}
\bm{v}_W \\  
\bm{q}_{WB} \cdot \mat{0 \\ \bm{\omega_B}/2} \\
\frac{1}{m}\;\bm{q}_{WB} \odot \bm{T}_B + \bm{g}_W \\
\bm{J}^{-1}\left(\boldsymbol{\tau}_B - \boldsymbol\omega_B \times \bm{J}\boldsymbol\omega_B\right)
\end{bmatrix}
\end{align}
with $\bm{g}_W= [0, 0, \SI{-9.81}{\meter\per\second^2}]^\intercal$ denoting Earth's gravity, $\bm{T}_B$ the collective thrust and $\bm{\tau}_B$ the body torque.
Again, an explicit \textit{Runge-Kutta} integration of 4th order is used.

{\bf Augmented Aerodynamic Residual Models.}
Following previous works \cite{Torrente2021Data-DrivenQuadrotors, Kabzan2019Learning-BasedRacing}, we use the data-driven model, in the form of a neural network $\mathcal{N}$, to complement the nominal dynamics by modeling a residual. In its full configuration, our residual dynamics model is defined as
\begin{align}\label{eq:full_dyn}
\begin{split}
\small
f(\bm{x}, \bm{u}) &=
f_{\mathcal{F}}(\bm{x}, \bm{u}) + f_{\mathcal{D}}(\bm{x}, \bm{u}) \; , \\
f_{\mathcal{D}}(\bm{x}, \bm{u}) &= 
\begin{bmatrix}
\bm{0}_2 \\  
f_{\mathcal{D}_\theta}(\bm{x}, \bm{u}) \\
f_{\mathcal{D}_\psi}(\bm{x}, \bm{u})
\end{bmatrix} \; ,
\end{split}
\end{align}
where we individually account for disturbances in linear and angular accelerations unknown to the nominal dynamics and $\theta$ and $\psi$ are the parameters of the neural networks modeling linear and angular disturbances respectively.

We also evaluate two simplified versions of the residual model:
\begin{align}\label{eq:part_dyn}
\small
\bm{f}_{\mathcal{D}_a}(\bm{x}, \bm{u}) = 
\begin{bmatrix}
\bm{0}_2 \\
\bm{f}_{\mathcal{D}_\theta}(\bm{v}_{B}) \\
0
\end{bmatrix},
\quad
\bm{f}_{\mathcal{D}_{a,u}}(\bm{x}, \bm{u}) = 
\begin{bmatrix}
\bm{0}_2 \\
\bm{f}_{\mathcal{D}_\theta}(\bm{v}_{B}, \bm{u}) \\
0
\end{bmatrix} \; .
\end{align}
These simplified models only consider residual forces as a function of the platform's velocity~(left), potentially accompanied by the commanded inputs~(right).

\newpage
{\bf Augmented Ground Effect Model.}\label{sec:method_groundeffect}
To show the strength of our approach, leveraging a complex arbitrary high level input, we extend the residual model using a height map under the quadrotor as additional input to model the ground effect.
\begin{align*}
\small
\bm{f}_{\mathcal{D}_{g}}(\bm{x}, \bm{u}) = 
\begin{bmatrix}
\bm{0}_2 \\
\bm{f}_{\mathcal{N}_\theta}(\bm{x}, \bm{u}, z_{WB} \cdot \bm{\mathds{1}} - h_{l}(\bm{p}_{WB}, \bm{H}_W)) \\
0
\end{bmatrix}
\end{align*}
where $z_{WB}$ is the altitude of the quadrotor and $h_{l}$ is a mapping $h_{l}: \mathbb{R}^{3} \times \mathbb{R}^{N \times M} \rightarrow \mathbb{R}^{3 \times 3}$ which takes the quadrotor's position $\bm{p}_{WB}$ and a fixed or sensed global height map $\bm{H}_W$ of size $N \times M$ as input. The function returns a $3 \times 3$ local patch of the height map around the quadrotor's position with a resolution of \SI{10}{cm}.

{\bf MPC Cost Formulation.}
We specify the cost in \cref{eq:nlp} to be of quadratic form $\mathcal{L}(\bm{x}, \bm{u}) = \| \bm{x} - \bm{x}_r  \|^2_Q + \| \bm{u} - \bm{u}_r \|^2_R$ penalizing deviations from a reference trajectory $\bm{x}_r, \bm{u}_r$ and account for input limitations by constraining $0 \leq \bm{u} \leq u_\text{max}$.

\section{Experiments}\label{sec:experiments}
In our experiments we will re-validate the findings of previous works \cite{Chee2022KNODE-MPC:Robots, Saviolo2022Physics-InspiredTracking} that using neural-network data-driven models in MPC improves tracking performance compared to no data-driven models or Gaussian Processes. More importantly, however, we will demonstrate that \emphalgname{} enables the use of larger network capacities to fully exhaust possible performance gains while providing real-time capabilities.

All our experiments are divided into two phases: system identification and evaluation. 
During system identification, we collect data using the nominal dynamics model in the MPC controller. 
The state-control-timeseries are further processed in subsequent state, control tuples. Each step is then re-simulated using the nominal controller and the error is used as the training label for the residual model.

During evaluation we track two fixed evaluation trajectories, Circle and Lemniscate, and measure the performance based on the reference position tracking error. 
As such, we report the (Mean) Euclidean Distance between the reference trajectory and the tracked trajectory as error. 

To identify model architectures used in the experiments we use a naming convention stating the model type followed by the size and the implementation type where we differentiate between our \emphalgname{} approach (-Ours) and a naive integration (-Naive). N-3-32-Ours is a neural network model with 3 hidden layers, 32 neurons each using our \emphalgname{} framework and N-3-32-Naive using a naive integration. GP-20 is a Gaussian Process Model with 20 inducing points.

All of our learned dynamic models are trained with a batch size of 64 and a learning rate of $1e^{-4}$ using the Adam optimizer. We split all datasets into a training and validation part and train the models using early stopping on the validation set. Dataset sizes are 20k datapoints for the simple simulation environment, 200k for the BEM simulation environment, and we use the openly available dataset presented in \cite{Bauersfeld2021NeuroBEM:Model} with 1.8 million datapoints for the real-world experiment.

When comparing against GPs we follow the original implementation of \cite{Torrente2021Data-DrivenQuadrotors} for the $\bm{f}_{\mathcal{D}_a}$ model configuration where one single-input-single-output GP is trained per dimension. For the $\bm{f}_{\mathcal{D}_{a,u}}$ configuration, their implementation is extended to a multi-input-single-output GP per dimension.

\subsection{Simulation} 
We use two simulation environments featuring varying modeling accuracy and real-time requirements to compare against a non-augmented MPC controller, a naive integration of data-driven dynamics \cite{Spielberg2021NeuralFriction, Chee2022KNODE-MPC:Robots, Saviolo2022Physics-InspiredTracking}, and GPs~\cite{Torrente2021Data-DrivenQuadrotors} with respect to real-time capability and model capacity. 

\begin{table}
\centering
\fontsize{6}{6}\selectfont
\renewcommand\tabcolsep{4.0pt}
\caption{Results for the Simplified Simulation experiment. Our deep learning models outperform Gaussian Processes even using small models. For large models our \emphalgname{} framework (-Ours) allows real-time capability without optimization time increase compared to the naive integration (-Naive).}
\begin{tabular}{@{}l|ccccc|ccccc|c@{}}
\toprule
  & \multicolumn{10}{c|}{Error {[}mm{]}} &  \multicolumn{1}{c}{$t$ {[}ms{]}} \\ \midrule
Track  & \multicolumn{5}{c|}{Circle} & \multicolumn{5}{c|}{Lemniscate} & avg \\ \midrule
$v_\text{avg}$ {[}m/s{]}  &  2   &  4.5   &  7   &  9.5   &  12   &  2  &  4.5    &   7   &   9.5   &   12   &      avg       \\
$v_\text{max}$ {[}m/s{]} & 2.1 &  4.8 &  7.5  & 10.2 & 12.8   &   2.9 &  5.9  & 10.5 & 14.0 & 18.1 & avg \\ \midrule
Perfect & 1 & 0 & 1 & 1 & 2 & 0 & 1 & 3 & 11 & 28 & 1.0 \\
Nominal & 50 & 134 & 213 & 277 & 333 & 50 & 124 & 187 & 244 & 297 & 1.0 \\ \midrule
GP-20 & 22 & 31 & 28 & 29 & 35 & 34 & 33 & 35 & 33 & 50 & 2.9 \\
GP-50 & 23 & 37 & 39 & 42 & 40 & 27 & 37 & 39 & 41 & 52 & 4.5 \\
GP-100 & 21 & 20 & 26 & 31 & 30 & 19 & 21 & 25 & 27 & 44 & 7.2 \\
N-1-12-Naive & 33 & 33 & 35 & 34 & 33 & 28 & 32 & 33 & 32 & 48 & 1.6 \\
N-1-12-Ours & 33 & 33 & 35 & 34 & 33 & 28 & 32 & 32 & 32 & 48 & 1.8 \\
N-2-18-Naive & 21 & 22 & 27 & 31 & 30 & 19 & 23 & 27 & 30 & \bf43 & 2.1 \\
N-2-18-Ours & 21 & 22 & 27 & 31 & 30 & 19 & 23 & 27 & 30 & \bf43 & 2.1 \\
N-3-32-Naive & 13 & 17 & 22 & 27 & 26 & 14 & 19 & \bf26 & 28 & 45 & 8.2 \\
N-3-32-Ours & 13 & 17 & 22 & 27 & 26 & 14 & 19 & \bf26 & 28 & 46 & 2.2 \\
N-4-64-Naive & \bf10 & \bf14 & 18 & 23 & \bf23 & \bf11 & \bf17 & 27 & \bf27 & 45 & \textcolor{red}{35.9} \\
N-4-64-Ours & \bf10 & \bf14 & \bf17 & \bf22 & \bf23 & \bf11 & \bf17 & 27 & \bf27 & 46 & 2.3 \\
N-5-128-Naive & 13 & 18 & 22 & 28 & 29 & 16 & 19 & 31 & 29 & 47 & \textcolor{red}{178.7} \\
N-5-128-Ours & 13 & 18 & 22 & 28 & 29 & 16 & 19 & 31 & 29 & 48 & 3.2 \\
\end{tabular}
\label{tab:results_simplified_sim}
\end{table}

{\bf Simplified Quadrotor Simulation.} We use the simulation framework described in~\cite{Torrente2021Data-DrivenQuadrotors}, where perfect odometry measurements and ideal tracking of the commanded single rotor thrusts are assumed. 
Drag effects by the rotors and fuselage are simulated, as well as zero mean (${\sigma=0.005}$) constant Gaussian noise on forces and torques, and zero mean Gaussian noise on motor voltage signals with standard deviation proportional to the input magnitude ${\sigma=0.02\sqrt{u}}$. There are no run-time constraints as controller and simulator are run sequentially in simulated time.  
Using the simplified simulation, we analyze the predictive performance and run-time of our approach for varying network sizes and directly compare to the naive implementation and Gaussian Process approach. 
We constrain the residual model to linear accelerations $\bm{f}_{\mathcal{D}_a}$ to facilitate comparison with prior work~\cite{Torrente2021Data-DrivenQuadrotors}. 
To fairly evaluate the run-times of our full and distributed approach and considering the limited resources of embedded systems this experiment was performed on a single CPU core. 
The results are depicted in~\cref{tab:results_simplified_sim}. We also compare with a \textit{Nominal} model where no learned residuals are modeled in the dynamics function and we also compare with an oracle-like \textit{Perfect} model which uses the same dynamics equations as the simulation (excluding noise).
Neural networks which achieve accurate modeling performance on the simulated dynamics are integrated easily with real-time optimization times below 3ms using our approach while they have high optimization times (up to 36ms) when a naive integration approach is used. The local approximations described in \cref{sec:approx} do not negatively influence performance compared to a naive implementation. Furthermore, we demonstrate that such modeling performance is not reachable with a GP even when using a large number of supporting points.

\begin{figure}
    \centering
    \includegraphics[width=\linewidth]{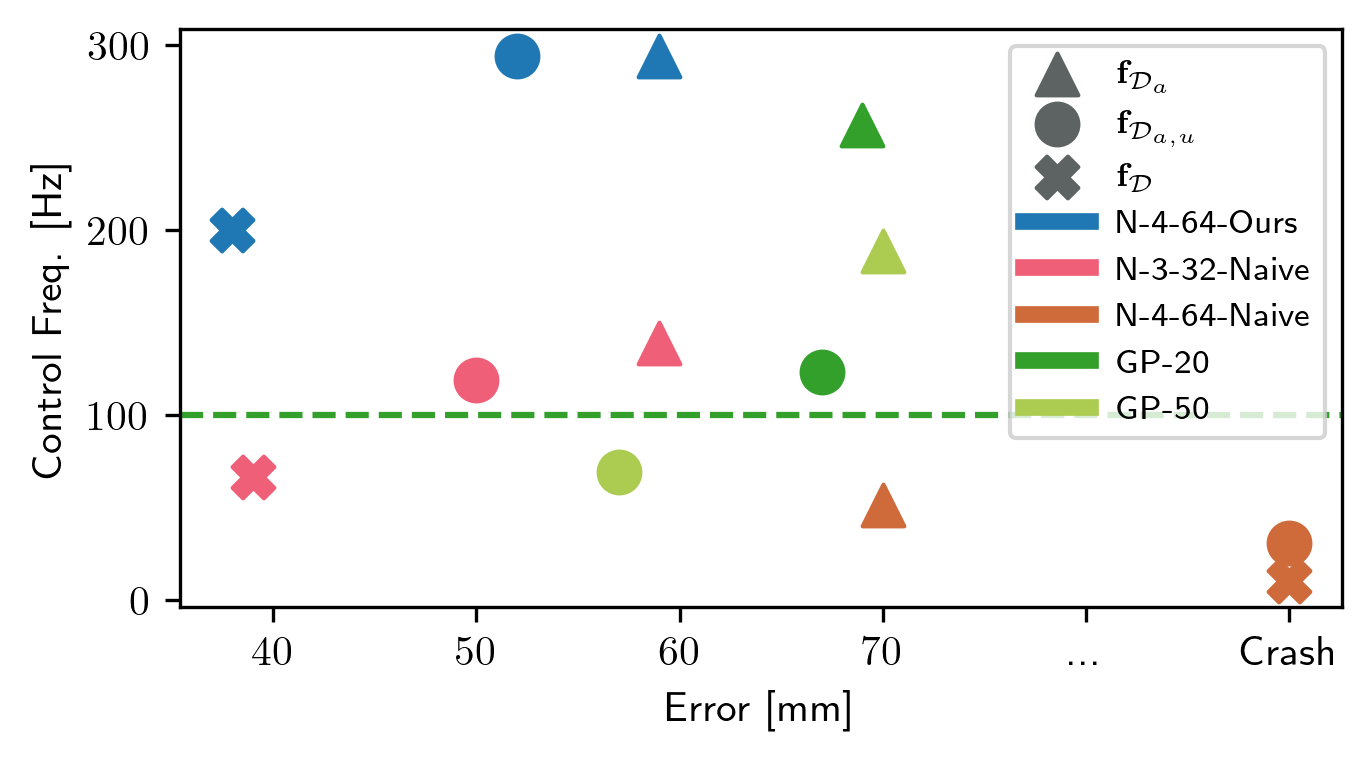}
    \caption{Control Frequency over Tracking Error for Lemniscate trajectory in the realistic BEM simulation - top-left is desired. Our approach (blue) can leverage multidimensional inputs and large model capacities while being real-time capable. Increasing the naive approach to a four layer network (orange) leads to the controller becoming unstable for high-dimensional input. No additional noise is simulated, leading to error standard deviations within $1mm$ over 5 trials per experiment, induced by non-deterministic ROS transportation times.}
    \label{fig:bem_res}
\end{figure}

{\bf BEM Quadrotor Simulation.} In addition to the simplified simulation setting, we also evaluate our approach in a highly accurate aerodynamics simulator based on Blade-Element-Momentum-Theory (BEM)~\cite{Bauersfeld2021NeuroBEM:Model}.
In contrast to the simplified simulation setting, this simulation can accurately model lift and drag produced by each rotor from the current ego-motion of the platform and the individual rotor speeds. The simulator runs in real-time and communicates with the controller via the Robot Operating System (ROS). We target a real-time control frequency of \SI{100}{Hz}.
We want to understand how our approach copes with increasing parameter count and model complexity of the learning task: First, we change the learned dynamics from just modeling linear acceleration residuals $\bm{f}_{\mathcal{D}_{a}}$ with velocity as inputs to also accounting for rotor commands $\bm{f}_{\mathcal{D}_{a,u}}$. In a second step we, model the full residual $\bm{f}_{\mathcal{D}}$ additionally outputting residuals on angular accelerations.
The results obtained in each of these settings are illustrated in \cref{fig:bem_res}. While a naive approach can accurately model the residuals, its control frequency quickly declines for increasingly complex models. For larger networks, it has excessively high optimization times leading to the controller becoming unstable even in simulation. In contrast, our \emphalgname{} approach can leverage both higher modeling capacity and the most representative residual model $\bm{f}_{\mathcal{D}}$ for on-par performance while running above 200Hz.

\subsection{Real World}\label{sec:experiments_real_world}

\begin{table}
\centering
\fontsize{7}{7}\selectfont
\caption{Results for the Real-World experiment. We improve tracking performance up to 82\% compared to the nominal controller and up to 55\% compared to GPs while being real-time capable unlike a naive integration. Result statistics are reported over 5 runs per experiment.} 
\begin{tabular}{@{}l|cc@{}}
\toprule
  & \multicolumn{2}{c}{Error {[}mm{]}} \\ \midrule
Track  & \multicolumn{1}{c}{Circle} & \multicolumn{1}{c}{Lemniscate}  \\ \midrule \midrule
$v_\text{max}$ {[}m/s{]} &  10   &  14        \\ \midrule
\cref{eq:nom_dyn}~$\bm{f}_{\mathcal{F}_{~}}$: \quad Nominal       &  321  &  359 \\ \midrule
\cref{eq:part_dyn}~$\bm{f}_{\mathcal{D}_a}$: \quad GP-20         &  66 $\pm$ 4  &  260 $\pm$ 11                       \\ 
\cref{eq:full_dyn}~$\bm{f}_{\mathcal{D}_{~}}$\hspace{0.4em}: \quad N-3-32-Naive        &  crash  & crash                        \\ 
\cref{eq:full_dyn}~$\bm{f}_{\mathcal{D}_{~}}$\hspace{0.4em}: \quad  N-3-32-Ours &  59 $\pm$ 6  & 117 $\pm$ 9                        \\
\end{tabular}
\label{tab:results_real_world}
\end{table}

Finally, we perform experiments evaluating the real world effectiveness of our approach by performing a set of agile trajectories using the physical quadrotor platform \textit{agilicious}~\cite{Foehn2022Agilicious:Flight}.
Control commands in the form of desired collective thrust and body rates are computed on a Jetson Xavier NX and are tracked by a low-level PID controller.
All real-world flight experiments are performed in an instrumented tracking arena that provides accurate pose estimates at $\SI{400}{\hertz}$.
As in the simulation experiments, we compare the tracking error along both circle and lemniscate trajectories at speeds up to $\SI{14}{\meter\per\second}$.
We evaluate our approach against the nominal controller, the naive integration, and the Gaussian Process configuration deployed in~\cite{Torrente2021Data-DrivenQuadrotors}. 
The results of these experiments is depicted in \cref{tab:results_real_world}, where we improve positional tracking error by up to 82\% compared to the nominal controller while the naive integration becomes unstable due to a long optimization time.
Furthermore, we outperform Gaussian Processes by up to 55\%. 

\begin{figure}
    \centering
    \begin{subfigure}{\linewidth}
        \centering
        \hspace{2.0em}
        \frame{\includegraphics[width=0.75\linewidth]{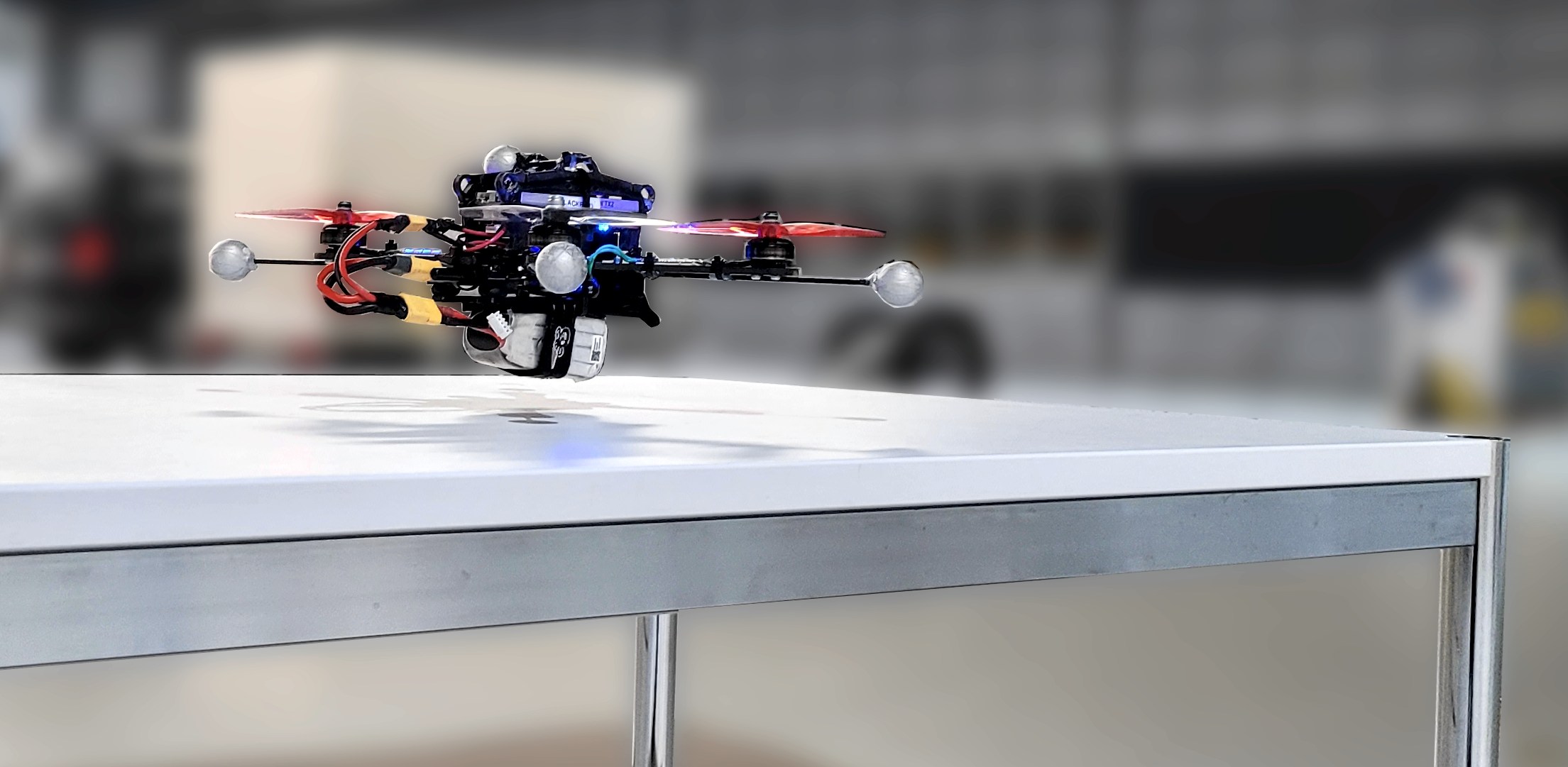}}
        \caption{}
    \end{subfigure}\vspace{0.5em}\\
    \begin{subfigure}{\linewidth}
        \centering
        \resizebox{0.85\linewidth}{!}{\input{resources/ground_effect}}
        \caption{}
    \end{subfigure}
    \vspace{-0.2em}
    \caption{(a) Quadrotor overflying the table in close proximity to the plane. (b) Vertical position error over distance. 
    Vertical lines mark the position of the table. 
    Our approach can model the aerodynamic effects in close proximity to the ground, substantially limiting the tracking error in z.}\vspace{-0.8em}
    \label{fig:ground_effect}
\end{figure}
{\bf Ground Effect.} 

Finally, we demonstrate the generalizability of our approach to other use-cases, modeling the complex aerodynamics of the ground effect using a height map as input (See \cref{sec:method_groundeffect}). 
We place a table of \SI{70}{cm} height in the flight arena and collect data by repeatedly flying over the table in close proximity. 
During evaluation, we fly repeated trajectories over the table with a target altitude of \SI{80}{cm} of the quadrotor's center of gravity; leaving approximately \SI{2}{cm} between the table and the lowest point of the quadrotor (battery). 
To isolate the performance of our approach, compensating for ground effect, we evaluate the trained model in two configurations. 
First, in which the height map information is unknown to the model (Baseline), and second where the information is known to the model. 
On an evaluation trajectory with 8 flyovers we improve the tracking error in z direction by 72\% in close proximity (table plane +\SI{10}{cm} in $xy$) above of the table. 
A visualization of a single flyover can be seen in \cref{fig:ground_effect}. 

\section{Conclusion}
In this work we demonstrated an approach to scale the modeling capacity of data-driven MPC using neural networks to larger, more powerful architectures while being real-time capable on embedded devices. Our framework can improve new and existing applications of data-driven MPC by increasing the available real-time modeling capacity; making our approach generalizable to a variety of control applications. 

\changes{An open challenge, which is not yet considered in this work, but the authors plan to tackle in the future, is to use a historic sequence of states and control input in a learned dynamics model. This would naturally lead to incorporating sequential and temporal models such (LSTMs, GRUs, and TCNs) in the optimization loop using our approach and would give rise to running approaches currently only feasible in simulation \cite{Bauersfeld2021NeuroBEM:Model} in embedded MPC real-time.} 

We experimentally show that the controller's performance is not negatively affected by the real-time inducing approximations. Thus, this method overcomes the limitation of having to sacrifice performance for efficiency as described in previous works \cite{Saviolo2022Physics-InspiredTracking, Chee2022KNODE-MPC:Robots}.
We demonstrate its usefulness by evaluating the isolated real-time capability of \emphalgname{} on different devices and applying the framework to the challenging problem of trajectory tracking of a highly agile quadrotor; reducing the tracking error substantially while using powerful models on-device.

\section*{Acknowledgment}
We thank Matteo Zallio for his help in visually communicating our work.
{
\printbibliography
}
\end{document}